# Optimal Variable Selection in Multi-Group Sparse Discriminant Analysis

Irina Gaynanova* and Mladen Kolar†

**Abstract**

This article considers the problem of multi-group classification in the setting where the number of variables $p$ is larger than the number of observations $n$. Several methods have been proposed in the literature that address this problem, however their variable selection performance is either unknown or suboptimal to the results known in the two-group case. In this work we provide sharp conditions for the consistent recovery of relevant variables in the multi-group case using the discriminant analysis proposal of Gaynanova et al. [7]. We achieve the rates of convergence that attain the optimal scaling of the sample size $n$, number of variables $p$ and the sparsity level $s$. These rates are significantly faster than the best known results in the multi-group case. Moreover, they coincide with the optimal minimax rates for the two-group case. We validate our theoretical results with numerical analysis.

## 1 Introduction

We consider a problem of multi-group classification in the high-dimensional setting, where the number of variables $p$ is much larger than the number of observations $n$. Given $n$ independent observations $\{(X_i, Y_i), i = 1, \ldots, n\}$ from a joint distribution $(X, Y)$ on $\mathbb{R}^p \times \{1, \ldots, G\}$, our goal is to learn a rule that will classify a new data point $X \in \mathbb{R}^p$ into one of the $G$ groups.

In a low dimensional setting (when $p \ll n$), Fisher's Linear Discriminant Analysis (FLDA) is a classical approach for obtaining a classification rule in the multi-group setting. To describe the FLDA, we introduce some additional notation. Denote $n_g$ the number of samples from the group $g$, $n_g = |\{i \mid Y_i = g\}|$, and the sample average in the group $g$ as $\bar{X}_g = n_g^{-1} \sum_{i|Y_i=g} X_i$. Let $W$ be a pooled sample covariance matrix,

$$W = (n - G)^{-1} \sum_{g=1}^{G} (n_g - 1) S_g, \tag{1.1}$$

where $S_g = (n_g - 1)^{-1} \sum_{i|Y_i=g}(X_i - \bar{X}_g)(X_i - \bar{X}_g)$. Furthermore, let $D = [D_1, \ldots, D_{G-1}] \in \mathbb{R}^{p \times (G-1)}$ be the matrix of sample mean contrasts between $G$ groups, with

$$D_r = \frac{\sqrt{n_{r+1}} \sum_{g=1}^{r} n_g (\bar{X}_g - \bar{X}_{r+1})}{\sqrt{n} \sqrt{\sum_{g=1}^{r} n_g \sum_{g=1}^{r+1} n_g}}, \quad r = 1, \ldots, G-1. \tag{1.2}$$

*Email: ig93@cornell.edu. Mailing address: Cornell University, Department of Statistical Science, 1173 Comstock Hall, Ithaca, NY 14853

†Email: mladen.kolar@chicagobooth.edu. Mailing address: The University of Chicago, Booth School of Business, 5807 Woodlawn Avenue, Chicago, IL 60637

FLDA estimates vectors $\{v_g\}_{g=1}^{G-1}$, which are linear combinations of $p$ variables, through the following optimization program

$$\begin{aligned} v_g = \arg\max_{v \in \mathbb{R}^p} & \left\{v^\top DD^\top v\right\} \\ \text{s.t.} \quad & v^\top W v = 1; \\ & v^\top W v_{g'} = 0 \quad \text{for} \quad g' < g. \end{aligned} \quad (1.3)$$

These combinations are called canonical vectors and they define the $(G-1)$-dimensional eigenspace of the matrix $W^{-1}DD^\top$ (see, for example, Chapter 11.5 of [13]). Given the matrix $V \in R^{p \times (G-1)}$ of vectors $\{v_g\}_{g=1}^{G-1}$, a new data point $X \in \mathbb{R}^p$ is classified into group $\widehat{g}$ if

$$\widehat{g} = \arg\min_{g \in \{1,\ldots,G\}} (X - \bar{X}_g)^\top V (V^\top W V)^{-1} V^\top (X - \bar{X}_g) - 2\log\frac{n_g}{n}. \quad (1.4)$$

This rule is a sample version of the optimal classification rule derived under the assumption of multivariate Gaussian class-conditional distributions with a common covariance matrix [14, Chapter 3.9]. Throughout the paper, we will assume that $X \mid Y = g \sim \mathcal{N}(\mu_g, \Sigma)$.

Unfortunately when the number of samples is small compared to the number of variables, the classification rule described above does not perform well [1, 17]. As a result a large body of literature has emerged to deal with classification in high-dimensions. To prevent overfitting, these methods assume that the optimal classification rule depends only on the few $s$ variables out of $p$. In the context of classification rule (1.4), this means that the matrix of canonical vectors $V$ only uses $s$ of these variables, that is, $V$ is row-sparse. In the context of binary classification, that is, when the number of groups is equal to 2, we point the reader to [17, 18, 21, 5, 11, 12, 23, 6, 3, 9] and references therein for recent progress on high-dimensional classification. Work on multi-group classification is less abundant. Initial progress has been reported in [4, 22], however, theoretical properties of the proposed methods were not studied. In a recent work, Gaynanova et al. [7] propose a convex estimation procedure that simultaneously estimates all the discriminant directions and establish sufficient conditions under which the correct set of discriminating variables is selected.

The focus of this paper is on establishing optimal conditions under which the Multi-Group Sparse Discriminant Analysis (MGSDA) procedure [7], described in §2, consistently recovers the relevant variables for classification. Consistent variable selection is an important property, since many domain scientist use the selected variables for hypothesis generation, downstream analysis and scientific discovery. [7] established equivalence between MGSDA and sparse discriminant analysis [12] in the two group case and then extended the proof technique of [12] to the multi-group case. This strategy, however, does not lead to optimal sample size scaling for consistent variable selection in the two group case [9]. In this paper, we use a refined proof strategy that allows us to establish consistent variable selection in the multi-group (with $G = \mathcal{O}(1)$) case under the same sample size scaling as in the two-group case. In particular, we establish that the sample size $n$ needs to satisfy

$$n \geq K \|\Sigma_{AA}^{-1}\|_2 \left(\max_{j \in A^c} \sigma_{jj \cdot A}\right)(G-1)s\log((p-s)\log(n))$$

in order for MGSDA to recover the correct variables. Here $K$ is a fixed constant independent from $n$, $p$, $s$ and $G$, and $\sigma_{jj \cdot A} = \Sigma_{jj} - \Sigma_{jA}\Sigma_{AA}^{-1}\Sigma_{Aj}$. At a high-level, we will follow the primal-dual strategy used in [9], however, there are a number of details that require careful dealing in order to establish the desired scaling. In particular, [7] showed that the solution to (1.3) is matrix $V = W^{-1}DR$, where $R$ is a $(G-1)$-dimensional orthogonal matrix. Furthermore, at the optima $\{v_g\}_{g=1}^{G-1}$, the objective values in (1.3) are equal to the non-zero eigenvalues of $D^\top W^{-1} D$. However, [7] separately considers the deviations of $W^{-1}$ and $D$ from their population counterparts, which is not sufficient to establish the optimal scaling of $(n, p, s)$ for consistent variable selection. In contrast, here we consider these quantities jointly. In the two-group case, $W^{-1}D$ is a vector and $D^\top W^{-1}D$ is a scalar, which allows [9] to use concentration inequalities for $\chi^2$ distributed random variables to achieve the optimal rate. In the multi-group case, one needs to characterize the joint distribution of the columns of $W^{-1}D$ and the behavior of the $\|D^\top W^{-1}D\|_2$, hence an analysis different from [9] is required. In particular, we use the distributional results of [2] to characterize $W^{-1}D$ and the results from random matrix theory [19, 20] for $\|D^\top W^{-1}D\|_2$.

The rest of the paper is organized as follows. In §2, we summarize the notation used throughout the paper and introduce the MGSDA procedure. In §3, we study the population version of the MGSDA estimator. Our main result is stated in §4. Illustrative simulation studies, which corroborate our theoretical findings, are provided in §5. Technical proofs are given in §7.

## 2 Preliminaries

In this section, we introduce the notation and the Multi-Group Sparse Discriminant Analysis problem.

For a vector $v \in \mathbb{R}^p$ we define $\|v\|_2 = \sqrt{\sum_{i=1}^p v_i^2}$, $\|v\|_1 = \sum_{i=1}^p |v_i|$, $\|v\|_\infty = \max_i |v_i|$. We use $e_j$ to define a unit norm vector with $j$th element being equal to 1. For a matrix $M$ we define by $m_i$ the $i$th row of $M$ and by $M_j$ the $j$th column of $M$. We also define $\|M\|_{\infty,2} = \max_i \|m_i\|_2$, $\|M\|_\infty = \|M\|_{\infty,\infty} = \max_i \|m_i\|_1$, $\|M\|_2 = \sigma_{\max}(M)$ and $\|M\|_F = \sqrt{\sum_i \sum_j m_{ij}^2}$. Given an index set $A$, we define $M_{AA}$ to be the submatrix of $M$ with rows and columns indexed by $A$. For two sequences $\{a_n\}$ and $\{b_n\}$, we write $a_n = \mathcal{O}(b_n)$ to define $a_n < Cb_n$ for some positive constant $C$. We write $a_n = o(b_n)$ to define $a_n b_n^{-1} \to 0$.

The MGSDA estimator [7] is found as the solution to the following convex optimization problem

$$\widehat{V} = \arg\min_{V \in \mathbb{R}^{p \times (G-1)}} \left\{ \frac{1}{2}\operatorname{Tr}(V^\top WV) + \frac{1}{2}\|D^\top V - I\|_F^2 + \lambda \sum_{i=1}^p \|v_i\|_2 \right\}, \tag{2.1}$$

where $W$ and $D$ are defined in (1.1) and (1.2), respectively. The sparsity of the estimated canonical vectors $\widehat{V}$ is controlled by the user specified parameter $\lambda > 0$. Note that the $\ell_2$-norm penalty encourages the rows of $\widehat{V}$ to be sparse leading to the variable selection. When $\lambda = 0$ and $W$ is nonsingular, $\widehat{V} = (W + DD^\top)^{-1}D$ spans the $(G-1)$-dimensional eigenspace of $W^{-1}DD^t$. Since the classification rule (1.4) is invariant with respect to linear transformations, the MGSDA coincides with classical sample canonical correlation analysis. Intuitively, the three components of the objective function in (2.1) minimize the within-class variability, control the level of between-class variability and provide regularization by inducing sparsity respectively.

In the next two sections, we study conditions under which the MGSDA consistently recovers the correct set of discriminant variables.

## 3 Variable Selection in the Population Setting

In this section, we develop understanding of the MGSDA in the limit of infinite amount of data. We will develop understanding of limitations of the procedure for the purpose of consistent variable selection.

Let $\pi_g$ be the prior group probabilities, $P(Y_i = g) = \pi_g$. Let $\mu_g$ be the population within-group mean, $\mu_g = \mathbb{E}(X_i \mid Y_i = g)$. Let $\Sigma$ be the population within-group covariance matrix, $\operatorname{Cov}(X_i \mid Y_i = g) = \Sigma$, and $\Delta \in \mathbb{R}^{p \times (G-1)}$ be the matrix of population mean contrasts between $G$ groups with $r$th column

$$\Delta_r = \frac{\sqrt{\pi_{r+1}} \sum_{g=1}^r \pi_g(\mu_g - \mu_{r+1})}{\sqrt{\sum_{g=1}^r \pi_g \sum_{g=1}^{r+1} \pi_g}}.$$

The population canonical vectors are eigenvectors of matrix $\Sigma^{-1}\Delta\Delta^\top$. The column vectors of matrix $\Psi = \Sigma^{-1}\Delta$ define the $(G-1)$-dimensional eigenspace of $\Sigma^{-1}\Delta\Delta^t$ [7]. Since the canonical vectors determine the variables that are relevant for the classification rule, in the high-dimensional setting we assume that the matrix $\Psi$ is row sparse. Let $A$ be the support of $\Psi$, $A = \{i \mid \|\Psi_i\|_2 \neq 0\}$, and $s$ be the cardinality of $A$, $s = |A|$.

The population version of MGSDA optimization problem is

$$\widehat{\Psi} = \arg\min_{V \in \mathbb{R}^{p \times (G-1)}} \left\{ \frac{1}{2} \operatorname{Tr}(V^\top \Sigma V) + \frac{1}{2} \|V^\top \Delta - I\|_F^2 + \lambda \sum_{i=1}^{p} \|v_i\|_2 \right\}. \quad (3.1)$$

Compared to the optimization program in (2.1), in (3.1) we assume access to the population covariance $\Sigma$ and mean contrasts $\Delta$. Theorem 1 characterizes conditions under which $\widehat{\Psi} = (\widehat{\Psi}_A^\top, 0_{p-s}^\top)^\top$ and $\|e_j^\top \widehat{\Psi}_A\|_2 \neq 0$ for all $j \in A$.

**Theorem 1.** *Suppose that*

$$\|\Sigma_{A^c A} \Sigma_{AA}^{-1} s_A\|_{\infty,2} < 1 \quad (3.2)$$

*and the tuning parameter $\lambda$ in (3.1) satisfies*

$$\lambda < \frac{\Psi_{\min}}{\|(\Sigma_{AA} + \Delta_A \Delta_A^\top)^{-1}\|_\infty \left(1 + \|\Delta_A^\top \Sigma_{AA}^{-1} \Delta_A\|_2\right)}, \quad (3.3)$$

*where $\Psi_{\min} = \min_{j \in A} \|e_j^\top \Psi_A\|_2 = \min_{j \in A} \|e_j^\top \Sigma^{-1} \Delta\|_2$. Then the solution $\widehat{\Psi}$ to (3.1) is of the form $\widehat{\Psi} = (\widehat{\Psi}_A^\top, 0_{p-s}^\top)^\top$, where*

$$\widehat{\Psi}_A = \Psi_A (I + \Delta_A^\top \Sigma_{AA}^{-1} \Delta_A)^{-1} - \lambda (\Sigma_{AA} + \Delta_A \Delta_A^\top)^{-1} s_A, \quad (3.4)$$

*and $s_A$ is the sub-gradient of $\sum_{i \in A} \|\widehat{\psi}_i\|_2$. Furthermore, we have that $\|e_j^\top \widehat{\Psi}_A\|_2 \neq 0$ for all $j \in A$.*

Theorem 1 provides sufficient conditions (3.2) and (3.3) under which the solution to (3.1) recovers the true support $A$. The condition (3.2) is of the same form as the irrepresentable condition in a multi-task regression [16]. The condition (3.3) relates the tuning parameter $\lambda$ and the minimal signal strength $\Psi_{\min}$. The tuning parameter $\lambda$ should not be too large, so that the relevant variables in $A$ are not shrank to zero. The upper bound depends on the minimal signal strength $\Psi_{\min}$ and the classification difficulty characterized by $\|\Delta_A^\top \Sigma_{AA}^{-1} \Delta_A\|_2$. Note that $\|(\Sigma_{AA} + \Delta_A \Delta_A^\top)^{-1}\|_\infty \leq \sqrt{s} \|(\Sigma_{AA} + \Delta_A \Delta_A^\top)^{-1}\|_2$, therefore it is sufficient for $\lambda$ to satisfy

$$\lambda < \frac{\Psi_{\min}}{\sqrt{s} \|(\Sigma_{AA} + \Delta_A \Delta_A^\top)^{-1}\|_2 \left(1 + \|\Delta_A^\top \Sigma_{AA}^{-1} \Delta_A\|_2\right)}. \quad (3.5)$$

Equation (3.4) provides an explicit form for the solution $\widehat{\Psi}$. Note that it estimates $\Psi_A$ up to the linear transformation $(I + \Delta_A^\top \Sigma_{AA}^{-1} \Delta_A)^{-1}$ and the bias term due to the penalty. The linear transformation has no effect on the support or the classification assignment due to invariance of classification rule (1.4). The bias term has no effect on the support as long as $\lambda$ satisfies (3.3). Note that Theorem 1 of [9] is a special case of our result in the two-group case.

## 4 Consistent Variable Selection of MGSDA

In this section, we establish our main result on the sample complexity needed for the variable selection consistency of the MGSDA.

We require the following assumptions.

**(C1)** Irrepresentability. There exists a constant $\alpha \in (0, 1]$ such that

$$\|\Sigma_{A^c A} \Sigma_{AA}^{-1} s_A\|_{\infty,2} \leq 1 - \alpha.$$

**(C2)** Minimal signal strength. There exists a constant $K_\psi > 0$ such that

$$\Psi_{\min} = \min_{j \in A} \|e_j^\top \Psi_A\|$$
$$\geq \lambda \sqrt{s} \|(\Sigma_{AA} + \Delta_A \Delta_A^\top)^{-1}\|_2 \times$$
$$\times \left(1 + K_\psi \left[\|\Delta_A^\top \Sigma_{AA}^{-1} \Delta_A\|_2 \vee 1\right] \left(1 + \sqrt{\max_{j \in A}(\Sigma_{AA}^{-1})_{jj} \frac{(G-1)\log(s\log(n))}{n}}\right)\right).$$

Irrepresentable condition is commonly used in the high-dimensional literature as a way to ensure exact variable selection of lasso like procedures [24, 20, 16, 9]. The second condition is commonly known as a beta-min condition and it states that the relevant variables should have sufficiently large signal in order for the procedure to distinguish them from noise.

Let $\widehat{A}$ be the support of $\widehat{V}$ defined in (2.1), $\widehat{A} = \{i : \|\widehat{v}_i\|_2 \neq 0\}$.

**Theorem 2.** *Assume that the conditions (C1) and (C2) are satisfied. Furthermore, suppose that the sample size satisfies*

$$n \geq K \left(\max_{j \in A^c} \sigma_{jj \cdot A}\right) \|\Sigma_{AA}^{-1}\|_2 (G-1) s \log((p-s)\log(n))$$

*for some absolute constant $K > 0$. If the tuning parameter $\lambda$ is selected as*

$$\lambda \geq K_\lambda (1 + \|\Delta_A^\top \Sigma_{AA}^{-1} \Delta_A\|_2)^{-1} \sqrt{\left(\max_{j \in A^c} \sigma_{jj \cdot A}\right) \frac{(G-1)\log((p-s)\log(n))}{n}},$$

*where $K_\lambda$ is an absolute constant that does not depend on the problem parameters, then the MGSDA procedure defined in (2.1) satisfies*

$$\widehat{A} = A,$$

*with probability at least $1 - \mathcal{O}(\log^{-1}(n))$.*

Theorem 2 is the finite sample version of Theorem 1. The main result states that the set of relevant variables will be recovered with high probability when the sample size $n$ is of the order $\mathcal{O}(s\log(p))$ and the minimal signal strength is of the order $\mathcal{O}\left(\sqrt{n^{-1}s\log(p)}\right)$. The $\sqrt{s}$ term in the minimal signal strength condition comes from the substitutions of $\|(\Sigma_{AA} + \Delta_A \Delta_A^\top)^{-1}\|_\infty$ by $\|(\Sigma_{AA} + \Delta_A \Delta_A^\top)^{-1}\|_2$. Theorem 2 significantly improves on the result in [7] which requires $n$ to be of the order $\mathcal{O}(s^2 \log(ps))$ and $\Psi_{\min}$ to be of the order $\mathcal{O}\left(\sqrt{n^{-1}s^2\log(ps)}\right)$. These improvements are achieved through the joint characterization of the distribution of $W_{AA}^{-1} D_A$ and deviations of $\|D_A^\top W_{AA}^{-1} D_A\|_2$ from $\|\Delta_A^\top \Sigma_{AA}^{-1} \Delta_A\|_2$. When $G = 2$, Theorem 2 reduces to the result established in [9] up to the condition on the tuning parameter $\lambda$. In [9] there is an additional factor $\sqrt{[\|\Delta_A^\top \Sigma_{AA}^{-1} \Delta_A\|_2 \vee 1]}$, which we avoid due to the use of a different proof technique.

### 4.1 Outline of the proof

The proof of Theorem 2 is based on the primal-dual witness technique [20]. In the course of the proof, one proposes a solution $\widehat{V}$ to (2.1) and verifies that the optimality conditions are satisfied.

We will verify that the vector $(\widetilde{V}_A^\top, 0^\top)^\top$, where $\widetilde{V}_A$ is the solution to the following oracle optimization program

$$\widetilde{V}_A = \arg\min_{V \in \mathbb{R}^{s \times (G-1)}} \frac{1}{2} \text{Tr}(V^\top W_{AA} V) + \frac{1}{2} \|D_A^\top V - I\|_F^2 + \lambda \sum_{i \in A} \|v_i\|_2,$$

satisfies the Karush-Kuhn-Tucker conditions for (2.1). The next lemma characterizes the form of the oracle solution $\widetilde{V}_A$.

**Lemma 3.** *The oracle solution satisfies*

$$\widetilde{V}_A = W_{AA}^{-1} D_A (I + D_A^\top W_{AA}^{-1} D_A)^{-1} - \lambda (W_{AA} + D_A D_A^\top)^{-1} s_A,$$

where $s_A$ is sub-gradient of $\sum_{i \in A} \|\widetilde{v}_i\|_2$.

Lemma 4 provides the sufficient conditions for the estimator $(\widetilde{V}_A^\top, 0^\top)^\top$ to be the oracle solution.

**Lemma 4.** *If*

$$\|(W_{A^c A} + D_{A^c} D_A^\top) \widetilde{V}_A - D_{A^c}\|_{\infty, 2} \leq \lambda; \tag{4.1}$$

$$\min_{j \in A} \|e_j^\top W_{AA}^{-1} D_A\|_2 > \lambda \|(W_{AA} + D_A D_A^\top)^{-1}\|_\infty (1 + \|D_A^\top W_{AA}^{-1} D_A\|_2), \tag{4.2}$$

then $\widehat{V} = (\widetilde{V}_A^\top, 0^\top)^\top$ and $\|e_j^\top \widetilde{V}_A\|_2 \neq 0$ for all $j \in A$.

Lemma 4 is deterministic in nature. We proceed to show that (4.1) and (4.2) are satisfied with high probability under conditions of Theorem 2. In particular, next theorem established that the correct variables $j$, $j \in A$, are estimated as nonzero by $\widetilde{V}_A$

**Theorem 5.** *Under conditions of Theorem 2, with probability at least $1 - \mathcal{O}(\log^{-1}(n))$*

$$\min_{j \in A} \|e_j^\top W_{AA}^{-1} D_A\|_2 > \lambda \|(W_{AA} + D_A D_A^\top)^{-1}\|_\infty (1 + \|D_A^\top W_{AA}^{-1} D_A\|_2).$$

To complete the proof, in the following theorem we establish that the wrong variables $j$, $j \in A^c$, are zero in $\widehat{V}$.

**Theorem 6.** *Under conditions of Theorem 2, with probability at least $1 - \mathcal{O}(\log^{-1}(n))$*

$$\|(W_{A^c A} + D_{A^c} D_A^\top) \widetilde{V}_A - D_{A^c}\|_{\infty, 2} \leq \lambda.$$

## 5 Simulation Results

We conduct several simulations to numerically illustrate finite sample properties of the MGSDA for the task of variable selection. The number of groups $G = 3$ and we change the size of the set $A$, $s \in \{10, 20, 30\}$, and the ambient dimension $p \in \{100, 200, 300\}$. The sample size is set as $n = \theta s \log(p)$ where $\theta$ is a control parameter that is varied. We report how well the MGSDA estimator recovers the set of variables $A$ as the control parameter $\theta$ varies. According to Theorem 2, the MGSDA recovers the correct variables when $n = K s \log(p)$ for some $K > 0$ and this will be illustrated in our simulations.

Next, we describe the data generating model. We set $\mathbb{P}(Y = g) = \frac{1}{3}$ for $g \in \{1, 2, 3\}$ and $X \mid Y = g \sim \mathcal{N}(\mu_g, \Sigma)$ with

$$\mu_1 = 0, \quad \mu_2 = (\underbrace{1, \ldots, 1}_{s}, \underbrace{0, \ldots, 0}_{p-s})^\top \quad \text{and} \quad \mu_3 = (\underbrace{1, \ldots, 1}_{s/2}, \underbrace{-1, \ldots, -1}_{s/2}, \underbrace{0, \ldots, 0}_{p-s})^\top.$$

We specify the covariance matrix $\Sigma$ as

$$\Sigma = \begin{pmatrix} \Sigma_{AA} & 0_{s \times p-s} \\ 0_{p-s \times s} & I_{p-s} \end{pmatrix}$$

and consider two cases for the component $\Sigma_{AA}$:

1. Toeplitz matrix, where $\Sigma_{TT} = [\Sigma_{ab}]_{a,b \in T}$ and $\Sigma_{ab} = \rho^{|a-b|}$ with $\rho \in \{0, 0.25, 0.5, 0.75, 0.9\}$, and
2. equal correlation matrix, where $\Sigma_{ab} = \rho$ when $a \neq b$ and $\sigma_{aa} = 1$, $\rho \in \{0, 0.25, 0.5, 0.75, 0.9\}$.

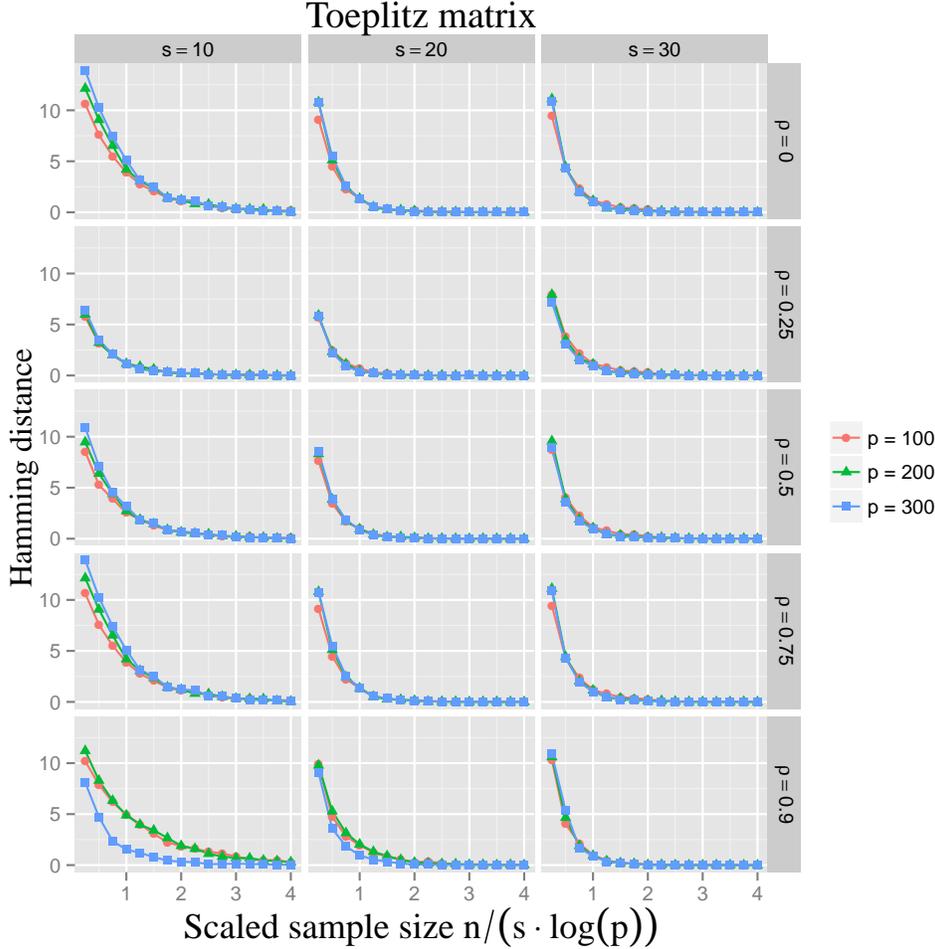

Figure 1: Performance of the MGSDA estimator averaged over 100 simulation runs. Plots of the rescaled sample size $n/(s\log(p))$ versus the Hamming distance between $\widehat{A}$ and $A$ for the Toeplitz matrix (see main text for details). Columns correspond to the size of $A$, $s \in \{10, 20, 30\}$, and rows correspond to different correlation strengths $\rho \in \{0, 0.25, 0.5, 0.75, 0.9\}$. Each subfigure shows three curves, corresponding to the problem sizes $p \in \{100, 200, 300\}$.

Finally, we set the penalty parameter as

$$\lambda = 0.5 \times \left(1 + \|\Delta_A^\top \Sigma_{AA}^{-1} \Delta_A^\top\|_2\right)^{-1} \sqrt{\frac{\log(p-s)}{n}}$$

for all cases, as suggested by Theorem 2. For each setting, we report the Hamming distance between the estimated set $\widehat{A}$ and the true set $A$ averaged over 200 independent simulation runs.

Figure 1 and Figure 2 illustrate finite sample performance of the MGSDA procedure. The Hamming distance is plotted against the control parameter $\theta$, which represents the rescaled number of samples. Each figure contains a number of subfigures, which correspond to different simulation settings. Columns correspond to different number of relevant variables, $|A| = s \in \{10, 20, 30\}$, and rows correspond to different values of $\rho$, $\rho \in \{0, 0.25, 0.5, 0.75, 0.9\}$. Each subfigure contains three curves for different problem sizes $p \in \{100, 200, 300\}$. We observe that as the control parameter $\theta$ increases the MGSDA procedure starts to recover the true set of variables, $A$, irrespective of the problem size, therefore, illustrating that our theoretical results describe well the finite sample performance of the procedure.

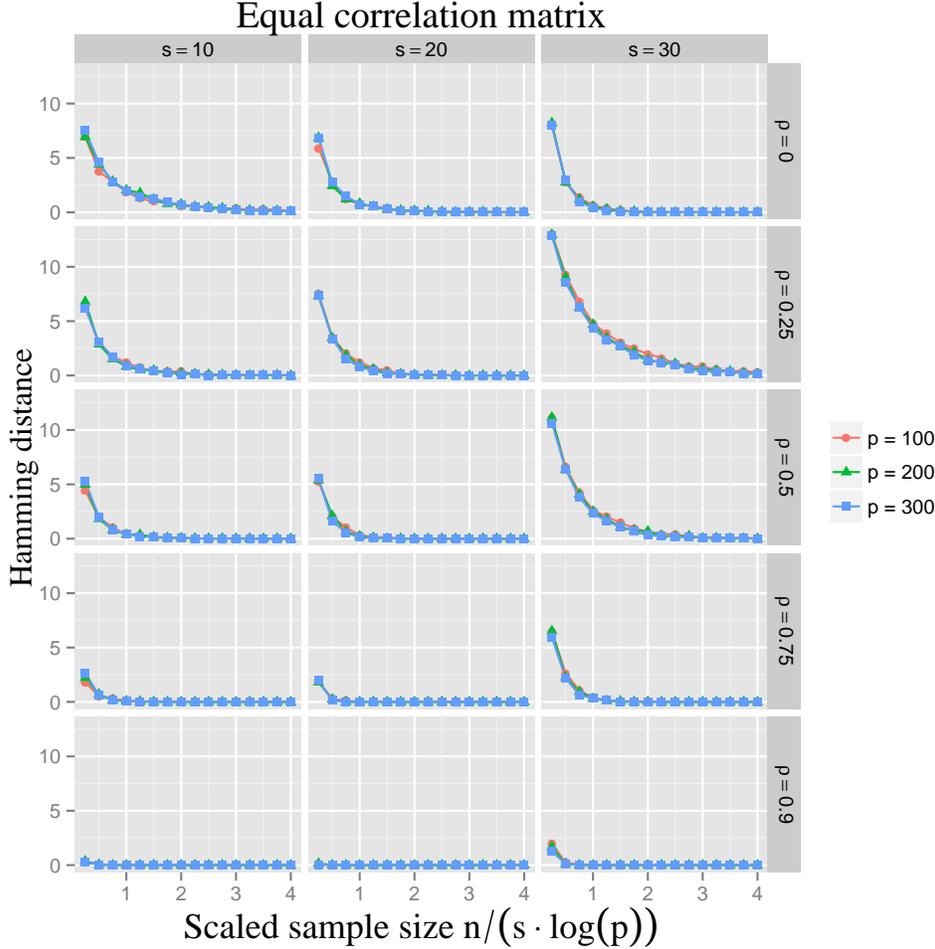

Figure 2: Performance of the MGSDA estimator averaged over 100 simulation runs. Plots of the rescaled sample size $n/(s\log(p))$ versus the Hamming distance between $\widehat{A}$ and $A$ for equal correlation matrix (see main text for details). Columns correspond to the size of $A$, $s \in \{10, 20, 30\}$, and rows correspond to different correlation strengths $\rho \in \{0, 0.25, 0.5, 0.75, 0.9\}$. Each subfigure shows three curves, corresponding to the problem sizes $p \in \{100, 200, 300\}$.

## 6   Discussion

In this paper we consider the problem of variable selection in discriminant analysis. This is the first time that the consistent variable selection in the multi-class setting has been established under the same conditions as in the two-class setting. Throughout the paper we have assumed that the number of classes $G$ does not increase with the sample size $n$, however this condition is not necessary for consistent variable selection and is used for the simplicity of exposition. We hope to address this issue in future work.

# 7 Technical Proofs

## 7.1 Proof of Theorem 1

Using the Karush-Kuhn-Tucker conditions, we have that any solution $\widehat{\Psi}$ of (3.1) satisfies

$$(\Sigma_{AA} + \Delta_A \Delta_A^\top)\widehat{\Psi}_A + (\Sigma_{AA^c} + \Delta_A \Delta_{A^c}^\top)\widehat{\Psi}_{A^c} - \Delta_A = -\lambda s_A; \tag{7.1}$$

$$(\Sigma_{A^c A} + \Delta_{A^c}\Delta_A^\top)\widehat{\Psi}_A + (\Sigma_{A^c A^c} + \Delta_{A^c}\Delta_{A^c}^\top)\widehat{\Psi}_{A^c} - \Delta_{A^c} = -\lambda s_{A^c}. \tag{7.2}$$

We proceed to verify that these conditions are satisfied by $\widehat{\Psi} = (\widehat{\Psi}_A^\top, 0^\top)^\top$ where $\widehat{\Psi}_A$ is given in (3.4). It is immediately clear that (7.1) is satisfied. We proceed to show that (7.2) is also satisfied. In particular, we show that

$$\|(\Sigma_{A^c A} + \Delta_{A^c}\Delta_A^\top)\widehat{\Psi}_A - \Delta_{A^c}\|_{\infty,2} < \lambda.$$

Since $\Sigma\Sigma^{-1}\Delta = \Delta$, it follows that $\Sigma_{A^c A}\Sigma_{AA}^{-1}\Delta_A = \Delta_{A^c}$. Therefore,

$$\begin{aligned}
\left(\Sigma_{A^c A} + \Delta_{A^c}\Delta_A^\top\right)\widehat{\Psi}_A &= (\Sigma_{A^c A} + \Delta_{A^c}\Delta_A^\top)(\Psi_A(I + \Delta_A^\top \Sigma_{AA}^{-1}\Delta_A)^{-1} - \lambda(\Sigma_{AA} + \Delta_A\Delta_A^\top)^{-1}s_A) \\
&= \Sigma_{A^c A}\Sigma_{AA}^{-1}\Delta_A(I + \Delta_A^\top \Sigma_{AA}^{-1}\Delta_A)^{-1} + \Delta_{A^c}\Delta_A^\top \Sigma_{AA}^{-1}\Delta_A(I + \Delta_A^\top \Sigma_{AA}^{-1}\Delta_A)^{-1} \\
&\quad - \lambda\Sigma_{A^c A}(\Sigma_A + \Delta_A\Delta_A^\top)^{-1}s_A - \lambda\Delta_{A^c}\Delta_A^\top(\Sigma_A + \Delta_A\Delta_A^\top)^{-1}s_A \\
&= \Delta_{A^c}(I + \Delta_A^\top \Sigma_{AA}^{-1}\Delta_A)^{-1} + \Delta_{A^c}(I - (I + \Delta_A^\top \Sigma_{AA}^{-1}\Delta_A)^{-1}) \\
&\quad - \lambda\Sigma_{A^c A}(\Sigma_{AA}^{-1} - \Sigma_{AA}^{-1}\Delta_A(I + \Delta_A^\top \Sigma_{AA}^{-1}\Delta_A)^{-1}\Delta_A^\top \Sigma_{AA}^{-1})s_A \\
&\quad - \lambda\Delta_{A^c}\Delta_A^\top(\Sigma_{AA}^{-1} - \Sigma_{AA}^{-1}\Delta_A(I + \Delta_A^\top \Sigma_{AA}^{-1}\Delta_A)^{-1}\Delta_A^\top \Sigma_{AA}^{-1})s_A \\
&= \Delta_{A^c} - \lambda\Sigma_{A^c A}\Sigma_{AA}^{-1}s_A + \lambda\Delta_{A^c}(I + \Delta_A^\top \Sigma_{AA}^{-1}\Delta_A)^{-1}\Delta_A^\top \Sigma_{AA}^{-1}s_A \\
&\quad - \lambda\Delta_{A^c}\Delta_A^\top \Sigma_{AA}^{-1}s_A + \lambda\Delta_{A^c}\Delta_A^\top \Sigma_{AA}^{-1}\Delta_A(I + \Delta_A^\top \Sigma_{AA}^{-1}\Delta_A)^{-1}\Delta_A^\top \Sigma_{AA}^{-1}s_A \\
&= \Delta_{A^c} - \lambda\Sigma_{A^c A}\Sigma_{AA}^{-1}s_A + \lambda\Delta_{A^c}\Delta_A^\top \Sigma_{AA}^{-1}s_A - \lambda\Delta_{A^c}\Delta_A^\top \Sigma_{AA}^{-1}s_A \\
&= \Delta_{A^c} - \lambda\Sigma_{A^c A}\Sigma_{AA}^{-1}s_A.
\end{aligned}$$

By assumption (3.2),

$$\|(\Sigma_{A^c A} + \Delta_{A^c}\Delta_A^\top)\widehat{\Psi}_A - \Delta_{A^c}\|_{\infty,2} = \lambda\|\Sigma_{A^c A}\Sigma_{AA}^{-1}s_A\|_{\infty,2} < \lambda,$$

which verifies that $\widehat{\Psi}$ also satisfies (7.2).

To complete the proof, we show that no component of $\widehat{\Psi}_A$ is set to zero. From (3.4),

$$e_j^\top \widehat{V}_A = e_j^\top \Psi_A(I + \Delta_A^\top \Sigma_{AA}^{-1}\Delta_A)^{-1} - \lambda e_j^\top (\Sigma_{AA} + \Delta_A\Delta_A^\top)^{-1}s_A.$$

Since

$$\|e_j^\top \Psi_A(I + \Delta_A^\top \Sigma_{AA}^{-1}\Delta_A)^{-1}\|_2 \geq \frac{1}{\|I + \Delta_A^\top \Sigma_{AA}^{-1}\Delta_A\|_2}\|e_j^\top \Psi_A\|_2 \geq \frac{\Psi_{\min}}{1 + \|\Delta_A^\top \Sigma_{AA}^{-1}\Delta_A\|_2}$$

and

$$\|\lambda e_j^\top (\Sigma_{AA} + \Delta_A\Delta_A^\top)^{-1}s_A\|_2 \leq \lambda\|(\Sigma_{AA} + \Delta_A\Delta_A^\top)^{-1}\|_\infty,$$

the result follows.

*Proof of Lemma 3 and 4.* The proof follows the proof of Theorem 1. □

*Proof of Theorem 5.* From Lemma 11, with probability at least $1 - \mathcal{O}(\log^{-1}(n))$

$$\|(W_{AA} + D_A D_A^\top)^{-1}\|_\infty \leq \sqrt{s}\|(\Sigma_{AA} + \Delta_A\Delta_A^\top)^{-1}\|_2\left(1 + \mathcal{O}\left(\sqrt{\frac{s\log(\log(n))}{n}}\right)\right).$$

From Lemma 14, with probability at least $1 - \mathcal{O}(\log^{-1}(n))$

$$\|D_A^\top W_{AA}^{-1} D_A\|_2 \leq C\|\Delta_A^\top \Sigma_{AA}^{-1} \Delta_A\|_2 + \mathcal{O}\left(\frac{(G-1)s\log(\log(n))}{n} \vee \sqrt{\|\Delta_A^\top \Sigma_{AA}^{-1} \Delta_A\|_2 \frac{(G-1)\log(\log(n))}{n}}\right).$$

Therefore, with probability at least $1 - \mathcal{O}(\log^{-1}(n))$

$$\lambda \|(W_{AA} + D_A D_A^\top)^{-1}\|_\infty (1 + \|D_A^\top W_{AA}^{-1} D_A\|_2)$$
$$\leq \lambda\sqrt{s}\|(\Sigma_{AA} + \Delta_A \Delta_A^\top)^{-1}\|_2 \left(1 + C\left[\|\Delta_A^\top \Sigma_{AA}^{-1} \Delta_A\|_2 \vee 1\right]\left(1 + \sqrt{\frac{(G-1)\log(\log(n))}{n}}\right)\right).$$

On the other hand, from Lemma 7, with probability at least $1 - \mathcal{O}(\log^{-1}(n))$

$$\min_{j \in A} \|e_j^\top W_{AA}^{-1} D_A\|_2$$
$$\geq \min_{j \in A} \|e_j^\top \Sigma_{AA}^{-1} \Delta_A\|_2 \left(1 - \mathcal{O}\left(\sqrt{\left[\|\Delta_A^\top \Sigma_{AA}^{-1} \Delta_A\|_2 \vee 1\right] \max_{j \in A}(\Sigma_{AA}^{-1})_{jj} \frac{(G-1)\log(s\log(n))}{n}}\right)\right)$$
$$\geq \Psi_{\min}\left(1 - \mathcal{O}\left(\sqrt{\left[\|\Delta_A^\top \Sigma_{AA}^{-1} \Delta_A\|_2 \vee 1\right] \max_{j \in A}(\Sigma_{AA}^{-1})_{jj} \frac{(G-1)\log(s\log(n))}{n}}\right)\right).$$

The final result follows from the condition on the sample size $n$ and (C2). $\square$

**Lemma 7.** *With probability at least $1 - \log^{-1}(n)$, $\forall j \in A$*

$$\|e_j^\top W_{AA}^{-1} D_A\|_2 \geq \|e_j^\top \Sigma_{AA}^{-1} \Delta_A\|_2 \left(1 - \mathcal{O}\left(\sqrt{\left[\|\Delta_A^\top \Sigma_{AA}^{-1} \Delta_A\|_2 \vee 1\right] (\Sigma_{AA}^{-1})_{jj} \frac{(G-1)\log(s\log(n))}{n}}\right)\right).$$

*Proof of Lemma 7.* By triangle inequality

$$\|e_j^\top W_{AA}^{-1} D_A - e_j^\top \Sigma_{AA}^{-1} \Delta_A\|_2 \leq \|e_j^\top W_{AA}^{-1} D_A - e_j^\top \Sigma_{AA}^{-1} D_A\|_2 + \|e_j^\top \Sigma_{AA}^{-1} D_A - e_j^\top \Sigma_{AA}^{-1} \Delta_A\|_2.$$

Consider the first term,

$$\|e_j^\top W_{AA}^{-1} D_A - e_j^\top \Sigma_{AA}^{-1} D_A\|_2$$
$$\leq e_j^\top W_{AA}^{-1} e_j \left\|\frac{D_A^\top W_{AA}^{-1} e_j}{e_j^\top W_{AA}^{-1} e_j} - \frac{D_A^\top \Sigma_{AA}^{-1} e_j}{e_j^\top \Sigma_{AA}^{-1} e_j}\right\|_2 + \|e_j^\top \Sigma_{AA}^{-1} D_A\|_2 \left|\frac{e_j^\top \Sigma_{AA}^{-1} e_j}{e_j^\top W_{AA}^{-1} e_j} - 1\right|.$$

From [9, Lemma 14], $\forall j \in A$

$$\left|\frac{e_j^\top \Sigma_{AA}^{-1} e_j}{e_j^\top W_{AA}^{-1} e_j} - 1\right| \leq C_2 \sqrt{\frac{\log(s\log(n))}{n}}$$

with probability at least $1 - (\log(n))^{-1}$. Further, using Lemma 10

$$\left\|\frac{D_A^\top W_{AA}^{-1} e_j}{e_j^\top W_{AA}^{-1} e_j} - \frac{D_A^\top \Sigma_{AA}^{-1} e_j}{e_j^\top \Sigma_{AA}^{-1} e_j}\right\|_2 = \|\widehat{H}_{12}\widehat{H}_{22}^{-1} - H_{12}H_{22}^{-1}\|_2 = \|\widehat{H}_{12}\widehat{H}_{22}^{-1} - \mu_h\|_2,$$

where

$$\widehat{H}_{12}\widehat{H}_{22}^{-1}|D_A \sim t_{G-1}(d_H, \mu_H, \Gamma_H)$$

with degrees of freedom $d_H = n - s - G + 2$, mean $\mu_H = H_{12}H_{22}^{-1}$ and scale parameter $\Gamma_H = \frac{1}{d_H}(D_A^\top R D_A)/(e_j^\top \Sigma_{AA}^{-1} e_j)$ with $R = \Sigma_{AA}^{-1} - \frac{\Sigma_{AA}^{-1} e_j e_j^\top \Sigma_{AA}^{-1}}{e_j^\top \Sigma_{AA}^{-1} e_j}$. Hence,

$$\widehat{H}_{12}\widehat{H}_{22}^{-1} - \mu_H = \frac{\Gamma_H^{1/2} y_H}{\sqrt{Z_H/d_H}} \quad \text{and} \quad \|\widehat{H}_{12}\widehat{H}_{22}^{-1} - \mu_H\|_2^2 = \frac{y_H^\top \Gamma_H y_H}{Z_H/d_H},$$

where $y_H \sim \mathcal{N}(0, I_{G-1})$ and $z_H \sim \chi_{d_H}^2$ are independent. Therefore,

$$P\left(\|\widehat{H}_{12}\widehat{H}_{22}^{-1} - \mu_H\|_2 \leq \sqrt{\frac{\epsilon_1}{\epsilon_2}}\right) = P\left(\|\widehat{H}_{12}\widehat{H}_{22}^{-1} - \mu_H\|_2^2 \leq \frac{\epsilon_1}{\epsilon_2}\right)$$
$$= P\left(\frac{y_H^\top \Gamma_H y_H}{Z_H/d_H} \leq \frac{\epsilon_1}{\epsilon_2}\right) \geq P(y_H^\top \Gamma_H y_H \leq \epsilon_1, Z_H/d_H \geq \epsilon_2)$$
$$\geq P(y_H^\top \Gamma_H y_H \leq \epsilon_1) P(Z_H/d_H \geq \epsilon_2).$$

Since $Z_H \sim \chi_{d_H}^2$, by Lemma 1 in [10] for all $y \geq 0$

$$P(Z_H/d_H \geq 1 - y) \geq 1 - \exp\left(-d_H \frac{y^2}{4}\right).$$

Since $y_H \sim \mathcal{N}(0, I_{G-1})$, using Proposition 1.1 in [8]

$$P(y_H^\top \Gamma_H y_H \geq \text{Tr}(\Gamma_H) + 2\sqrt{\text{Tr}(\Gamma_H^2)t} + 2\|\Gamma_H\|_2 t) \leq \exp(-t).$$

Combining the above displays,

$$\|\widehat{H}_{12}\widehat{H}_{22}^{-1} - \mu_H\|_2 \leq \sqrt{\frac{\text{Tr}(\Gamma_H) + 2\sqrt{\text{Tr}(\Gamma_H^2)t} + 2\|\Gamma_H\|_2 t}{1-y}}$$

with probability at least

$$(1 - \exp(-t))(1 - \exp(-d_H \frac{y^2}{4})) = 1 - (\exp(-t) + \exp(-d_H y^2/4) - \exp(-t)\exp(-d_H y^2/4)).$$

Setting it to be $1 - \mathcal{O}(\log^{-1}(n))$ for all $j \in A$, we get $t = \log(s\log(n))$, $y = 2\sqrt{\frac{\log(s\log(n))}{n-s-G+2}}$ and

$$\|\widehat{H}_{12}\widehat{H}_{22}^{-1} - \mu_H\|_2 \leq \sqrt{\frac{\text{Tr}(\Gamma_H) + 2\sqrt{\text{Tr}(\Gamma_H^2)\log(s\log(n))} + 2\|\Gamma_H\|_2 \log(s\log(n))}{1 - 2\sqrt{\frac{\log(s\log(n))}{n-s-G+2}}}}.$$

Since $\text{Tr}(\Gamma_H) \leq (G-1)\|\Gamma_H\|_2$ and $\text{Tr}(\Gamma_H^2) \leq (G-1)^2\|\Gamma_H\|_2^2$, the above display can be rewritten as

$$\|\widehat{H}_{12}\widehat{H}_{22}^{-1} - \mu_H\|_2$$
$$\leq \sqrt{\|\Gamma_H\|_2((G-1) + 2(G-1)\sqrt{\log(s\log(n))} + 2\log(\log(n))\left(1 + O\left(\sqrt{\frac{\log(s\log(n))}{n}}\right)\right)}.$$

Hence, there exists constant $C > 0$ such that with probability at least $1 - O(\log^{-1}(n))$

$$\|\widehat{H}_{12}\widehat{H}_{22}^{-1} - \mu_H\|_2 \leq C\sqrt{\|\Gamma_H\|_2(G-1)\log(s\log(n))}.$$

Using the definition of $R$,

$$\|\Gamma_H\|_2 = \frac{1}{n-s-G-2}\frac{1}{(\Sigma_{AA}^{-1})_{jj}}\|D_A^\top R D_A\|_2 \leq \frac{1}{n-s-G-2}\frac{1}{(\Sigma_{AA}^{-1})_{jj}}\|D_A^\top \Sigma_{AA}^{-1} D_A\|_2.$$

Applying Lemma 13, with probability at least $1 - \log^{-1}(n)$

$$\|\Gamma_H\|_2 \leq C\frac{1}{n-s-G-2}\frac{1}{(\Sigma_{AA}^{-1})_{jj}}\left[\|\Delta_A^\top \Sigma_{AA}^{-1}\Delta_A\|_2 \vee \sqrt{\|\Delta_A^\top \Sigma_{AA}^{-1}\Delta_A\|_2}\right].$$

Therefore, with probability at least $1 - \log^{-1}(n)$

$$\|\widehat{H}_{12}\widehat{H}_{22}^{-1} - \mu_H\|_2 \leq \mathcal{O}\left(\sqrt{\frac{\left[\|\Delta_A^\top \Sigma_{AA}^{-1}\Delta_A\|_2 \vee 1\right]}{(\Sigma_{AA}^{-1})_{jj}}\frac{(G-1)\log(s\log(n))}{n}}\right).$$

Consider $\|e_j^\top \Sigma_{AA}^{-1} D_A - e_j^\top \Sigma_{AA}^{-1}\Delta_A\|_2$. From Lemma 8, $\Sigma_{AA}^{-1}D_A \sim \mathcal{N}(\Sigma_{AA}^{-1}\Delta_A, \frac{\Sigma_{AA}^{-1}}{n}\otimes I_{G-1})$. Hence,

$$P\left(\|e_j^\top \Sigma_{AA}^{-1}D_A - e_j^\top \Sigma_{AA}^{-1}\Delta_A\|_2 \geq \epsilon\right) \leq P\left(\sqrt{G-1}\|e_j^\top \Sigma_{AA}^{-1}D_A - e_j^\top \Sigma_{AA}^{-1}\Delta_A\|_\infty \geq \epsilon\right)$$

$$\leq 2(G-1)\exp\left(-\frac{n\epsilon^2}{2(\Sigma_{AA}^{-1})_{jj}(G-1)}\right).$$

Let $\epsilon = \sqrt{2(\Sigma_{AA}^{-1})_{jj}(G-1)\frac{\log(2(G-1)s\log(n))}{n}}$. Then for all $j \in A$

$$\|e_j^\top \Sigma_{AA}^{-1}D_A - e_j^\top \Sigma_{AA}^{-1}\Delta_A\|_2 \leq \sqrt{2(\Sigma_{AA}^{-1})_{jj}(G-1)\frac{\log(2(G-1)s\log(n))}{n}}$$

with probability at least $1 - \log^{-1}(n)$. Also,

$$\|e_j^\top \Sigma_{AA}^{-1}D_A\|_2 \leq \|e_j^\top \Sigma_{AA}^{-1}D_A - e_j^\top \Sigma_{AA}^{-1}\Delta_A\|_2 + \|e_j^\top \Sigma_{AA}^{-1}\Delta_A\|_2$$

$$\leq \|e_j^\top \Sigma_{AA}^{-1}\Delta_A\|_2 + \sqrt{2(\Sigma_{AA}^{-1})_{jj}(G-1)\frac{\log(2(G-1)s\log(n))}{n}}.$$

Combining the above displays, with probability at least $1 - (\log(n))^{-1}$, for all $j \in A$

$$\|e_j^\top W_{AA}^{-1}D_A - e_j^\top \Sigma_{AA}^{-1}\Delta_A\|_2$$

$$\leq C_1(\Sigma_{AA}^{-1})_{jj}\sqrt{\frac{\left[\|\Delta_A^\top \Sigma_{AA}^{-1}\Delta_A\|_2 \vee 1\right]}{(\Sigma_{AA}^{-1})_{jj}}\frac{(G-1)\log(s\log(n))}{n}}$$

$$+ \|e_j^\top \Sigma_{AA}^{-1}\Delta_A\|_2 C_2\sqrt{\frac{\log(s\log(n))}{n}} + C_3\sqrt{(\Sigma_{AA}^{-1})_{jj}(G-1)\frac{\log(s\log(n))}{n}}$$

$$\leq C\|e_j^\top \Sigma_{AA}^{-1}\Delta_A\|_2\sqrt{\left[\|\Delta_A^\top \Sigma_{AA}^{-1}\Delta_A\|_2 \vee 1\right](\Sigma_{AA}^{-1})_{jj}\frac{(G-1)\log(s\log(n))}{n}}.$$

The final result follows form triangle inequality. □

*Proof of Theorem 6.* Since $(n-G)W \sim W_p(n-G, \Sigma)$, then $(n-G)W = UU^\top$, where $U \in \mathbb{R}^{p\times(n-G)}$ with columns $u_i \stackrel{iid}{\sim} \mathcal{N}(0, \Sigma)$. Let

$$E_D = D_{A^c} - \Sigma_{A^c A}\Sigma_{AA}^{-1}D_A;$$
$$E_U = U_{A^c} - \Sigma_{A^c A}\Sigma_{AA}^{-1}U_A \text{ with } (n-G)W_{A^c A} = U_{A^c}U_A^\top.$$

Then,
$$D_{A^c} = \Sigma_{A^c A}\Sigma_{AA}^{-1}D_A + E_D;$$
$$(n-G)W_{A^c A} = U_{A^c}U_A^\top = (\Sigma_{A^c A}\Sigma_{AA}^{-1}U_A + E_U)U_A^\top = \Sigma_{A^c A}\Sigma_{AA}^{-1}(n-G)W_{AA} + E_U U_A^\top.$$

and therefore

$$
\begin{aligned}
&(W_{A^c A} + D_{A^c}D_A^\top)\widetilde{V}_A - D_{A^c}\\
=&(\Sigma_{A^c A}\Sigma_{AA}^{-1}W_{AA} + (n-G)^{-1}E_U U_A^\top\\
&+ (\Sigma_{A^c A}\Sigma_{AA}^{-1}D_A + E_D)D_A^\top)(W_{AA}^{-1}D_A(I + D_A^\top W_{AA}^{-1}D_A)^{-1} - \lambda(W_{AA} + D_A D_A^\top)^{-1}s_A)\\
&- \Sigma_{A^c A}\Sigma_{AA}^{-1}D_A - E_D\\
=&\Sigma_{A^c A}\Sigma_{AA}^{-1}(D_A(I + D_A^\top W_{AA}^{-1}D_A)^{-1} + D_A D_A^\top W_{AA}^{-1}D_A(I + D_A^\top W_{AA}^{-1}D_A)^{-1} - D_A)\\
&+ \Sigma_{A^c A}\Sigma_{AA}^{-1}(-\lambda W_{AA}(W_{AA} + D_A D_A^\top)^{-1}s_A - \lambda D_A D_A^\top(W_{AA} + D_A D_A^\top)^{-1}s_A)\\
&+ (n-G)^{-1}E_U U_A^\top(W_{AA}^{-1}D_A(I + D_A^\top W_{AA}^{-1}D_A)^{-1} - \lambda(W_{AA} + D_A D_A^\top)^{-1}s_A)\\
&+ E_D(D_A^\top W_{AA}^{-1}D_A(I + D_A^\top W_{AA}^{-1}D_A)^{-1} - \lambda D_A^\top(W_{AA} + D_A D_A^\top)^{-1}s_A - I)\\
=& -\lambda\Sigma_{A^c A}\Sigma_{AA}^{-1}s_A + (n-G)^{-1}E_U U_A^\top(W_{AA} + D_A D_A^\top)^{-1}(D_A - \lambda s_A)\\
&- E_D(\lambda D_A^\top(W_{AA} + D_A D_A^\top)^{-1}s_A + (I + D_A^\top W_{AA}^{-1}D_A)^{-1})\\
=& -\lambda\Sigma_{A^c A}\Sigma_{AA}^{-1}s_A + (n-G)^{-1}E_U U_A^\top(W_{AA} + D_A D_A^\top)^{-1}(D_A - \lambda s_A)\\
&- E_D(I + D_A^\top W_{AA}^{-1}D_A)^{-1}(\lambda D_A^\top W_{AA}^{-1}s_A + I)
\end{aligned}
$$

We would like to establish the following:

$$\lambda\|\Sigma_{A^c A}\Sigma_{AA}^{-1}s_A\|_{\infty,2} < \lambda(1-\alpha) \tag{7.3}$$
$$\|(n-G)^{-1}E_U U_A^\top W_{AA}^{-1}D_A(I + D_A^\top W_{AA}^{-1}D_A)^{-1}\|_{\infty,2} < \lambda\alpha/4 \tag{7.4}$$
$$\lambda\|(n-G)^{-1}E_U U_A^\top W_{AA}^{-1}(I + W_{AA}^{-1}D_A D_A^\top)^{-1}s_A\|_{\infty,2} \leq \lambda\alpha/4 \tag{7.5}$$
$$\lambda\|E_D(I + D_A^\top W_{AA}^{-1}D_A)^{-1}D_A^\top W_{AA}^{-1}s_A\|_{\infty,2} \leq \lambda\alpha/4 \tag{7.6}$$
$$\|E_D(I + D_A^\top W_{AA}^{-1}D_A)^{-1}\|_{\infty,2} \leq \lambda\alpha/4. \tag{7.7}$$

**1. Show** $\|E_D(I + D_A^\top W_{AA}^{-1}D_A)^{-1}\|_{\infty,2} \leq \lambda\alpha/4.$

Consider $E_D = \Sigma_{A^c A}\Sigma_{AA}^{-1}D_A - D_{A^c}$. Since $\Sigma\Sigma^{-1}\Delta = \Delta$, it follows that $\Sigma_{A^c A}\Sigma_{AA}^{-1}\Delta_A = \Delta_{A^c}$. Hence $\mathbb{E}(D_{A^c}) = \Delta_{A^c} = \Sigma_{A^c A}\Sigma_{AA}^{-1}\Delta_A$. Therefore $\mathbb{E}(E_D) = 0$. Moreover,

$$
\begin{aligned}
\operatorname{Cov}(E_D, D_A) &= \operatorname{Cov}(\Sigma_{A^c A}\Sigma_{AA}^{-1}D_A - D_{A^c}, D_A) = \operatorname{Cov}(\Sigma_{A^c A}\Sigma_{AA}^{-1}D_A, D_A) - \operatorname{Cov}(D_{A^c}, D_A)\\
&= \Sigma_{A^c A}\Sigma_{AA}^{-1}\operatorname{Cov}(D_A) - \operatorname{Cov}(D_{A^c}, D_A) = \Sigma_{A^c A}\Sigma_{AA}^{-1}\operatorname{Cov}(D_A) - \Sigma_{A^c A}\Sigma_{AA}^{-1}\operatorname{Cov}(D_A)\\
&= 0.
\end{aligned}
$$

From Lemma 8, for all $j \in A^c$

$$e_j^\top E_D \sim \mathcal{N}\left(0, \frac{1}{n}\sigma_{jj \cdot A}I_{G-1}\right)$$

where $\sigma_{jj \cdot A} = \Sigma_{jj} - \Sigma_{jA}\Sigma_{AA}^{-1}\Sigma_{Aj}$ and $e_j^\top E_D$ is independent of $D_A$. Note that

$$
\begin{aligned}
\|E_D(I + D_A^\top W_{AA}^{-1}D_A)^{-1}\|_{\infty,2} &= \max_{j \in A^c}\|e_j^\top E_D(I + D_A^\top W_{AA}^{-1}D_A)^{-1}\|_2\\
&\leq \frac{\max_{j \in A^c}\|e_j^\top E_D\|_2}{1 + \sigma_{\min}(D_A^\top W_{AA}^{-1}D_A)}\\
&\leq \max_{j \in A^c}\|e_j^\top E_D\|_2
\end{aligned}
$$

Using Proposition 1.1 in [8]

$$\bigcap_{j \in A^C} \left\{ \frac{\|e_j^\top E_D\|_2^2}{\sigma_{jj \cdot A}} \leq \frac{(G-1)}{n} + 2\frac{\sqrt{(G-1)\log((p-s)\log(n))}}{n} + 2\frac{\log((p-s)\log(n))}{n} \right\}$$

with probability at least $1 - \log^{-1}(n)$. Hence, with probability at least $1 - \log^{-1}(n)$

$$\max_{j \in A^c} \frac{\|e_j^\top E_D\|_2^2}{\sigma_{jj \cdot A}} \leq \mathcal{O}\left(\frac{(G-1)\log((p-s)\log(n))}{n}\right),$$

or equivalently

$$\max_{j \in A^c} \|e_j^\top E_D\|_2 \leq \mathcal{O}\left(\sqrt{\max_{j \in A^c} \sigma_{jj \cdot A} \frac{(G-1)\log((p-s)\log(n))}{n}}\right).$$

**2. Show** $\lambda \|E_D(I + D_A^\top W_{AA}^{-1} D_A)^{-1} D_A^\top W_{AA}^{-1} s_A\|_{\infty,2} \leq \lambda \alpha/4$.
Since $e_j^\top E_D \sim \mathcal{N}\left(0, n^{-1}\sigma_{jj \cdot A} I_{G-1}\right)$, it follows that

$$e_j^\top E_D(I + D_A^\top W_{AA}^{-1} D_A)^{-1} D_A^\top W_{AA}^{-1} s_A \sim \mathcal{N}\left(0, \frac{\sigma_{jj \cdot A}}{n} s_A^\top W_{AA}^{-1} D_A(I + D_A^\top W_{AA}^{-1} D_A)^{-2} D_A^\top W_{AA}^{-1} s_A\right).$$

Following the above arguments, the following event has probability at least $1 - \log^{-1}(n)$

$$\bigcap_{j \in A^C} \left\{ \frac{\|e_j^\top E_D(I + D_A^\top W_{AA}^{-1} D_A)^{-1} D_A^\top W_{AA}^{-1} s_A L^{-1/2}\|_2^2}{\sigma_{jj \cdot A}} \leq \mathcal{O}\left(\frac{(G-1)\log((p-s)\log(n))}{n}\right) \right\},$$

where $L = s_A^\top W_{AA}^{-1} D_A(I + D_A^\top W_{AA}^{-1} D_A)^{-2} D_A^\top W_{AA}^{-1} s_A$. This implies that with probability at least $1 - \log^{-1}(n)$

$$\max_{j \in A^c} \frac{\|e_j^\top E_D(I + D_A^\top W_{AA}^{-1} D_A)^{-1} D_A^\top W_{AA}^{-1} s_A\|_2^2}{\sigma_{jj \cdot A}} \leq \|L\|_2 \mathcal{O}\left(\frac{(G-1)\log((p-s)\log(n))}{n}\right)$$

By triangle inequality

$$\begin{aligned}
\|L\|_2 &= \|s_A^\top W_{AA}^{-1} D_A(I + D_A^\top W_{AA}^{-1} D_A)^{-2} D_A^\top W_{AA}^{-1} s_A\|_2 \\
&\leq \|s_A^\top W_{AA}^{-1} s_A\|_2 \|(I + D_A^\top W_{AA}^{-1} D_A)^{-1} D_A^\top W_{AA}^{-1/2}\|_2^2 \\
&\leq \|s_A^\top W_{AA}^{-1} s_A\|_2 \\
&\leq s \|s_A\|_{\infty,2}^2 \|W_{AA}^{-1}\|_2 \\
&\leq s \|W_{AA}^{-1}\|_2.
\end{aligned}$$

From Lemma 9 in [20], with probability at least $1 - \log^{-1}(n)$

$$\|W_{AA}^{-1}\|_2 \leq \|\Sigma_{AA}^{-1}\|_2 \left(1 + \mathcal{O}\left(\sqrt{\frac{s \log(\log(n))}{n}}\right)\right).$$

Combining the above displays, with probability at least $1 - \mathcal{O}(\log^{-1}(n))$

$$\max_{j \in A^c} \frac{\|e_j^\top E_D(I + D_A^\top W_{AA}^{-1} D_A)^{-1} D_A^\top W_{AA}^{-1} s_A\|_2^2}{\sigma_{jj \cdot A}} \leq \|\Sigma_{AA}^{-1}\|_2 \mathcal{O}\left(\frac{(G-1)s \log((p-s)\log(n))}{n}\right),$$

or equivalently

$$\max_{j \in A^c} \|e_j^\top E_D (I + D_A^\top W_{AA}^{-1} D_A)^{-1} D_A^\top W_{AA}^{-1} s_A\|_2$$

$$\leq \mathcal{O}\left(\sqrt{\|\Sigma_{AA}^{-1}\|_2 \max_{j \in A^c} \sigma_{jj \cdot A} \frac{(G-1)s \log((p-s)\log(n))}{n}}\right).$$

**3. Show** $\|(n-G)^{-1} E_U U_A^\top W_{AA}^{-1} D_A (I + D_A^\top W_{AA}^{-1} D_A)^{-1}\|_{\infty,2} < \lambda \alpha/4$.

By definition $E_U = U_{A^c} - \Sigma_{A^c A} \Sigma_{AA}^{-1} U_A$, hence

$$\text{vec}(E_U) \sim \mathcal{N}(0, \Sigma_{A^c A^c \cdot A} \otimes I_{n-G})$$

and is independent of $U_A$. Therefore

$$(n-G)^{-1} e_j^\top E_U \sim \mathcal{N}\left(0, \frac{1}{(n-G)^2} \sigma_{jj \cdot A} I_{n-G}\right),$$

where $\sigma_{jj \cdot A} = \Sigma_{jj} - \Sigma_{jA} \Sigma_{AA}^{-1} \Sigma_{Aj}$. Conditional on $X_A$,

$$\frac{1}{n-G} e_j^\top E_U U_A^\top W_{AA}^{-1} D_A (I + D_A^\top W_{AA}^{-1} D_A)^{-1}$$

$$\sim \mathcal{N}\left(0, \frac{\sigma_{jj \cdot A}}{n-G} (I + D_A^\top W_{AA}^{-1} D_A)^{-1} D_A^\top W_{AA}^{-1} D_A (I + D_A^\top W_{AA}^{-1} D_A)^{-1}\right).$$

Let $(I + D_A^\top W_{AA}^{-1} D_A)^{-1} D_A^\top W_{AA}^{-1} D_A (I + D_A^\top W_{AA}^{-1} D_A)^{-1} = L$. Then by Proposition 1.1 in [8]

$$\bigcap_{j \in A^C} \left\{ \frac{\|(n-G)^{-1} e_j^\top E_U U_A^\top W_{AA}^{-1} D_A (I + D_A^\top W_{AA}^{-1} D_A)^{-1} L^{-1/2}\|_2^2}{\sigma_{jj \cdot A}} \right.$$

$$\left. \leq \frac{(G-1)}{n-G} + 2 \frac{\sqrt{(G-1)\log((p-s)\log(n))}}{n-G} + 2 \frac{\log((p-s)\log(n))}{n-G} \right\}$$

with probability at least $1 - \log^{-1}(n)$. Therefore,

$$\bigcap_{j \in A^C} \left\{ \frac{\|(n-G)^{-1} e_j^\top E_U U_A^\top W_{AA}^{-1} D_A (I + D_A^\top W_{AA}^{-1} D_A)^{-1}\|_2^2}{\sigma_{jj \cdot A}} \right.$$

$$\left. \leq \|L\|_2 \mathcal{O}\left(\frac{(G-1)\log((p-s)\log(n))}{n-G}\right) \right\}$$

with probability at least $1 - \log^{-1}(n)$. Since

$$\|L\|_2 = \|(I + D_A^\top W_{AA}^{-1} D_A)^{-1} D_A^\top W_{AA}^{-1} D_A (I + D_A^\top W_{AA}^{-1} D_A)^{-1}\|_2$$
$$= \|(I + D_A^\top W_{AA}^{-1} D_A)^{-2} D_A^\top W_{AA}^{-1} D_A\|_2 < 1,$$

with probability at least $1 - \log^{-1}(n)$

$$\max_{j \in A^c} \frac{\|(n-G)^{-1} e_j^\top E_U U_A^\top W_{AA}^{-1} D_A (I + D_A^\top W_{AA}^{-1} D_A)^{-1}\|_2^2}{\sigma_{jj \cdot A}} \leq \mathcal{O}\left(\frac{(G-1)\log((p-s)\log(n))}{n-G}\right),$$

or equivalently

$$\max_{j \in A^c} \|(n-G)^{-1} e_j^\top E_U U_A^\top W_{AA}^{-1} D_A (I + D_A^\top W_{AA}^{-1} D_A)^{-1}\|_2$$

$$\leq \mathcal{O}\left(\sqrt{\max_{j \in A^c} \sigma_{jj \cdot A} \frac{(G-1)\log((p-s)\log(n))}{n-G}}\right),$$

**4. Show** $\lambda \|(n-G)^{-1} E_U U_A^\top W_{AA}^{-1}(I + W_{AA}^{-1} D_A D_A^\top)^{-1} s_A\|_{\infty,2} \leq \lambda \alpha/4$.

Since $(n-G)^{-1} e_j^\top E_U \sim \mathcal{N}\left(0, (n-G)^{-2} \sigma_{jj \cdot A} I_{n-G}\right)$, it follows that

$$\frac{1}{n-G} e_j^\top E_U U_A^\top (W_{AA} + D_A D_A^\top)^{-1} s_A \sim \mathcal{N}\left(0, \frac{\sigma_{jj \cdot A}}{n-G} s_A^\top (W_{AA} + D_A D_A^\top)^{-1} W_{AA} (W_{AA} + D_A D_A^\top)^{-1} s_A\right).$$

Similar to parts 2 and 3, with probability at least $1 - \log^{-1}(n)$

$$\max_{j \in A^c} \|\frac{1}{n-G} e_j^\top E_U U_A^\top (W_{AA} + D_A D_A^\top)^{-1} s_A\|_2$$
$$\leq \mathcal{O}\left(\sqrt{\|L\|_2 \max_{j \in A^c} \sigma_{jj \cdot A} \frac{(G-1)\log((p-s)\log(n))}{n}}\right),$$

where

$$\|L\|_2 = \|s_A^\top (W_{AA} + D_A D_A^\top)^{-1} W_{AA} (W_{AA} + D_A D_A^\top)^{-1} s_A\|_2$$
$$= \|W_{AA}^{1/2} (W_{AA} + D_A D_A^\top)^{-1} s_A\|_2^2$$
$$\leq s \|W_{AA}^{1/2} W_{AA}^{-1/2} (I + W_{AA}^{-1/2} D_A D_A^\top W_{AA}^{-1/2})^{-1} W_{AA}^{-1/2}\|_2^2$$
$$\leq s \|(I + W_{AA}^{-1/2} D_A D_A^\top W_{AA}^{-1/2})^{-1}\|_2^2 \|W_{AA}^{-1/2}\|_2^2$$
$$\leq s \|W_{AA}^{-1}\|_2.$$

Following the same argument as in part 2, with probability at least $1 - \mathcal{O}(\log^{-1} n)$

$$\max_{j \in A^c} \|\frac{1}{n-G} e_j^\top E_U U_A^\top (W_{AA} + D_A D_A^\top)^{-1} s_A\|_2$$
$$\leq \mathcal{O}\left(\sqrt{\|\Sigma_{AA}^{-1}\|_2 \max_{j \in A^c} \sigma_{jj \cdot A} \frac{(G-1)s\log((p-s)\log(n))}{n}}\right).$$

**Combining 1-4**. The equations (7.4)-(7.7) are satisfied with probability at least $1 - \mathcal{O}(\log^{-1}(n))$ if for some constants $C_1 \geq 0$ and $C_2 \geq 0$

$$\alpha \geq C_1 \sqrt{\|\Sigma_{AA}^{-1}\|_2 \max_{j \in A^c} \sigma_{jj \cdot A} \frac{(G-1)s\log((p-s)\log(n))}{n}}$$

and

$$\lambda \geq \frac{1}{\alpha} C_2 \sqrt{\max_{j \in A^c} \sigma_{jj \cdot A} \frac{(G-1)\log((p-s)\log(n))}{n-G}}.$$

These inequalities are satisfied by (C1) and the conditions on sample size $n$ and tuning parameter $\lambda$ from Theorem 2. □

## 7.2 Auxillary Technical Results

**Lemma 8.** *If $X_i | Y_i = g \sim \mathcal{N}(\mu_g, \Sigma)$ for $i = 1, ..., n$, then*

$$D \sim \mathcal{N}(\Delta + o(1), \Sigma/n \otimes I + o(1)); \quad (n-G)W_p \sim W(\Sigma, n-G).$$

**Remark 9.** *The bias term $o(1)$ does not depend on either $s$ or $p$, and therefore we don't consider this term in the remaining analysis.*

*Proof of Lemma 8.* The result for $W$ is trivial. The definition of $D$ and the multivariate normality assumption on $X_i$ imply $D \sim \mathcal{N}(\mu_D, \Sigma_{D1} \otimes \Sigma_{D2})$. It remains to show $\mu_D = \Delta + o(1)$, $\Sigma_{D1} = \Sigma/n$ and $\Sigma_{D2} = I$. Consider the $r$th column of $D$,

$$D_r = \frac{\sqrt{n_{r+1}} \sum_{g=1}^{r} n_g(\bar{X}_g - \bar{X}_{r+1})}{\sqrt{n}\sqrt{\sum_{g=1}^{r} n_g \sum_{g=1}^{r+1} n_g}},$$

and the $r$th column of $\Delta$,

$$\Delta_r = \frac{\sqrt{\pi_{r+1}} \sum_{g=1}^{r} \pi_g(\mu_g - \mu_{r+1})}{\sqrt{\sum_{g=1}^{r} \pi_g \sum_{g=1}^{r+1} \pi_g}}.$$

Note that $\mathbb{E}(\bar{X}_i - \bar{X}_j) = \mu_i - \mu_j$ for all $i, j \in \{1, ..., G\}$. Moreover, $(n_1, ..., n_G) \sim Mult(n, (\pi_1, ..., \pi_G))$, and therefore $\mathbb{E}(n_i/n) = \pi_i$ and $\text{Cov}(n_i/n, n_j/n) = \pi_i \pi_j/n$ for all $i, j \in \{1, ..., G\}$. Hence,

$$\mathbb{E}(D_r) = \mathbb{E}(\mathbb{E}(D_r | n_1, ..., n_G)) = \mathbb{E}\left(\frac{\sqrt{n_{r+1}} \sum_{g=1}^{r} n_g \mathbb{E}((\bar{X}_g - \bar{X}_{r+1})|n_1,...,n_G)}{\sqrt{n}\sqrt{\sum_{g=1}^{r} n_g \sum_{g=1}^{r+1} n_g}}\right)$$

$$= \mathbb{E}\left(\frac{\sqrt{n_{r+1}} \sum_{g=1}^{r} n_g(\mu_g - \mu_{r+1})}{\sqrt{n}\sqrt{\sum_{g=1}^{r} n_g \sum_{g=1}^{r+1} n_g}}\right)$$

$$= \Delta_r + o(1).$$

First, consider the case $n_g/n = \pi_g$ for all $g \in \{1, ..., G\}$. Since the groups are independent,

$$\text{Cov}(D_r) = \mathbb{E}\left\{(D_r - \Delta_r)(D_r - \Delta_r)^\top\right\}$$

$$= \frac{1}{Gr(r+1)} \mathbb{E}\left\{\left(\sum_{i=1}^{r}(\bar{x}_i - \mu_i) - r(\bar{x}_{r+1} - \mu_{r+1})\right)\left(\sum_{i=1}^{r}(\bar{x}_i - \mu_i) - r(\bar{x}_{r+1} - \mu_{r+1})\right)^\top\right\}$$

$$= \frac{1}{Gr(r+1)}\left\{\sum_{i=1}^{r} \mathbb{E}\left\{(\bar{x}_i - \mu_i)(\bar{x}_i - \mu_i)^\top\right\} + r^2 \mathbb{E}\left\{(\bar{x}_{r+1} - \mu_{r+1})(\bar{x}_{r+1} - \mu_{r+1})^\top\right\}\right\}$$

$$= \frac{1}{Gr(r+1)}(r + r^2)\frac{\Sigma}{n/G} = \frac{\Sigma}{n},$$

and for $s > r$

$$\text{Cov}(D_r, D_s)$$
$$= \mathbb{E}\left\{(D_r - \Delta_r)(D_s - \Delta_s)^\top\right\}$$
$$= \frac{1}{G\sqrt{r(r+1)s(s+1)}} \mathbb{E}\left\{\left(\sum_{i=1}^{r}(\bar{x}_i - \mu_i) - r(\bar{x}_{r+1} - \mu_{r+1})\right)\left(\sum_{i=1}^{s}(\bar{x}_i - \mu_i) - s(\bar{x}_{s+1} - \mu_{s+1})\right)^\top\right\}$$
$$= \frac{1}{G\sqrt{r(r+1)s(s+1)}}\left\{\sum_{i=1}^{r} \mathbb{E}\left\{(\bar{x}_i - \mu_i)(\bar{x}_i - \mu_i)^\top\right\} - r\mathbb{E}\left\{(\bar{x}_{r+1} - \mu_{r+1})(\bar{x}_{r+1} - \mu_{r+1})^\top\right\}\right\}$$
$$= \frac{1}{G\sqrt{r(r+1)s(s+1)}}(r - r)\frac{\Sigma}{n/G} = 0.$$

The final result follows since $|n_i/n - \pi_i| = o(1)$. $\square$

**Lemma 10.**
$$\frac{D_A^\top W_{AA}^{-1} e_j}{e_j^\top W_{AA}^{-1} e_j} | D_A \sim t_{G-1}(d_H, \mu_H, \Gamma_H)$$

with degrees of freedom $d_H = n - s - G + 2$, mean $\mu_H = D_A^\top \Sigma_{AA}^{-1} e_j / (e_j^\top \Sigma_{AA}^{-1} e_j)$ and scale parameter $\Gamma_H = \frac{1}{d_H}(D_A^\top R D_A)/(e_j^\top \Sigma_{AA}^{-1} e_j)$ with $R = \Sigma_{AA}^{-1} - \frac{\Sigma_{AA}^{-1} e_j e_j^\top \Sigma_{AA}^{-1}}{e_j^\top \Sigma_{AA}^{-1} e_j}$.

*Proof of Lemma 10.* Let
$$H = \begin{pmatrix} D_A^\top \Sigma_{AA}^{-1} D_A & D_A^\top \Sigma_{AA}^{-1} e_j \\ e_j^\top \Sigma_{AA}^{-1} D_A & e_j^\top \Sigma_{AA}^{-1} e_j \end{pmatrix} = \begin{pmatrix} H_{11} & H_{12} \\ H_{12}^\top & H_{22} \end{pmatrix},$$

and

$$\widehat{H} = \begin{pmatrix} D_A^\top W_{AA}^{-1} D_A & D_A^\top W_{AA}^{-1} e_j \\ e_j^\top W_{AA}^{-1} D_A & e_j^\top W_{AA}^{-1} e_j \end{pmatrix} = \begin{pmatrix} \widehat{H}_{11} & \widehat{H}_{12} \\ \widehat{H}_{12}^\top & \widehat{H}_{22} \end{pmatrix}.$$

By definition, $\frac{D_A^\top W_{AA}^{-1} e_j}{e_j^\top W_{AA}^{-1} e_j} = \widehat{H}_{12} \widehat{H}_{22}^{-1}$. Let $M = (D_A\ e_j)^\top \in \mathbb{R}^{G \times s}$. Then $H$ can be rewritten as $H = M \Sigma_{AA}^{-1} M^\top$ and $\widehat{H}$ as $\widehat{H} = M W_{AA}^{-1} M^\top$. Since $(n - G) W_{AA} \sim W_s(n - G, \Sigma_{AA})$ and $\text{rank}(M) = G$, by [15, Theorem 3.2.11]

$$(n - G) \widehat{H}^{-1} \sim W_G(n - s, H^{-1}),$$

or equivalently

$$\frac{1}{n - G} \widehat{H} \sim W_G^{-1}(n - s + G + 1, H).$$

By definition of $R$, $H_{11 \cdot 2} = D_A^\top R D_A$. Using [2, Theorem 3], $\widehat{H}_{12} \widehat{H}_{22}^{-1}$ has density

$$f_{\widehat{H}_{12} \widehat{H}_{22}^{-1}}(X) = \frac{|D_A^\top R D_A|^{-\frac{1}{2}} |e_j^\top \Sigma_{AA}^{-1} e_j|^{\frac{G-1}{2}}}{\pi^{(G-1)/2}} \frac{\Gamma(\frac{n-s+1}{2})}{\Gamma(\frac{n-s-G+2}{2})}$$
$$\times |I + e_j^\top \Sigma_{AA}^{-1} e_j (D_A^\top R D_A)^{-1} (X - H_{12} H_{22}^{-1})(X - H_{12} H_{22}^{-1})^\top|^{-\frac{1}{2}(n-s+1)}.$$

Since $|I + uv^\top| = 1 + u^\top v$,

$$f_{\widehat{H}_{12} \widehat{H}_{22}^{-1}}(X) = \frac{|D_A^\top R D_A|^{-\frac{1}{2}} |e_j^\top \Sigma_{AA}^{-1} e_j|^{\frac{G-1}{2}}}{\pi^{(G-1)/2}} \frac{\Gamma(\frac{n-s+1}{2})}{\Gamma(\frac{n-s-G+2}{2})}$$
$$\times \left(1 + e_j^\top \Sigma_{AA}^{-1} e_j (X - H_{12} H_{22}^{-1})^\top (D_A^\top R D_A)^{-1} (X - H_{12} H_{22}^{-1})\right)^{-\frac{1}{2}(n-s+1)}.$$

This density corresponds to a $(G-1)$-dimensional elliptical $t$-distribution with $n - s - G + 2$ degrees of freedom, mean $\mathbb{E}(\widehat{H}_{12} \widehat{H}_{22}^{-1}) = H_{12} H_{22}^{-1}$ and $\text{Cov}(\widehat{H}_{12} \widehat{H}_{22}^{-1}) = \frac{1}{n-s-G} \frac{D_A^\top R D_A}{e_j^\top \Sigma_{AA}^{-1} e_j}$. □

**Lemma 11.** *With probability at least $1 - \mathcal{O}(\log^{-1}(n))$*

$$\|(W_{AA} + D_A D_A^\top)^{-1}\|_\infty \leq \sqrt{s} \|(\Sigma_{AA} + \Delta_A \Delta_A^\top)^{-1}\|_2 \left(1 + \mathcal{O}\left(\sqrt{\frac{s \log(\log(n))}{n}}\right)\right).$$

*Proof of Lemma 11.* First, we prove that unconditional distribution of $X_{Ai} \in \mathbb{R}^s$, $i = 1, ..., n$, is sub-gaussian: for all $x \in \mathbb{R}^s$, $< X_{Ai}, x >$ is sub-gaussian. Since $X_{Ai} | Y_i = g \sim \mathcal{N}(\mu_{gA}, \Sigma_{AA})$, $X_{Ai}$ can be expressed as

$$X_{Ai} = C_{Ai} + Z_{Ai},$$

where $Z_{Ai} \sim \mathcal{N}(0, \Sigma_{AA})$ and $P(C_{Ai} = \mu_{gA}) = \pi_g$ for $g = 1, ..., G$. Let $\widetilde{x} = <X_{Ai}, x>$, $\widetilde{c} = <C_{Ai}, x>$ and $\widetilde{z} = <Z_{Ai}, x>$. Then $\widetilde{x} = \widetilde{c} + \widetilde{z}$. Consider the sub-gaussian norm of $\widetilde{x}$ [19, Definition 5.7]

$$\|\widetilde{x}\|_{\psi_2} = \sup_{d \geq 1} d^{-1/2} \left(\mathbb{E}|\widetilde{x}|^d\right)^{1/d}.$$

By triangle inequality, $\|\widetilde{x}\|_{\psi_2} \leq \|\widetilde{c}\|_{\psi_2} + \|\widetilde{z}\|_{\psi_2}$. Note that $\|\widetilde{c}\|_{\psi_2}$ is finite for all $x$ since $C_{Ai}$ is a bounded random vector, and $\|\widetilde{z}\|_{\psi_2}$ is finite for all $x$ since $Z_{Ai}$ is a zero-mean gaussian random vector. It follows that $\|\widetilde{x}\|_{\psi_2}$ is finite for all $x$, hence $X_{Ai}$ is unconditionally sub-gaussian.

By definition, $\Sigma_{AA} + \Delta_A \Delta_A^\top$ is unconditional population covariance matrix of $X_A$ and $W_{AA} + D_A D_A^\top$ is unconditional sample covariance matrix of $X_A$. Using Theorem 5.39 in [19], with probability at least $1 - \log^{-1}(n)$

$$\|(\Sigma_{AA} + \Delta_A \Delta_A^\top)^{-1/2} (W_{AA} + D_A D_A^\top)(\Sigma_{AA} + \Delta_A \Delta_A^\top)^{-1/2} - I\|_2 \leq C \sqrt{\frac{s \log(\log(n))}{n}}.$$

By submultiplicity of operator norm,

$$\|(W_{AA} + D_A D_A^\top)^{-1} - (\Sigma_{AA} + \Delta_A \Delta_A^\top)^{-1}\|_2$$
$$\leq \|(\Sigma_{AA} + \Delta_A \Delta_A^\top)^{-1}\|_2 \|(\Sigma_{AA} + \Delta_A \Delta_A^\top)^{1/2} (W_{AA} + D_A D_A^\top)^{-1} (\Sigma_{AA} + \Delta_A \Delta_A^\top)^{1/2} - I\|_2.$$

Therefore, with probability at least $1 - \log^{-1}(n)$

$$\|(W_{AA} + D_A D_A^\top)^{-1} - (\Sigma_{AA} + \Delta_A \Delta_A^\top)^{-1}\|_2 \leq C \|(\Sigma_{AA} + \Delta_A \Delta_A^\top)^{-1}\|_2 \sqrt{\frac{s \log(\log(n))}{n}}.$$

By triangle inequality,

$$\|(W_{AA} + D_A D_A^\top)^{-1}\|_\infty \leq \|(\Sigma_{AA} + \Delta_A \Delta_A^\top)^{-1}\|_\infty + \|(\Sigma_{AA} + \Delta_A \Delta_A^\top)^{-1} - (W_{AA} + D_A D_A^\top)^{-1}\|_\infty$$
$$\leq \sqrt{s} \|(\Sigma_{AA} + \Delta_A \Delta_A^\top)^{-1}\|_2 + \sqrt{s} \|(\Sigma_{AA} + \Delta_A \Delta_A^\top)^{-1} - (W_{AA} + D_A D_A^\top)^{-1}\|_2$$
$$\leq \sqrt{s} \|(\Sigma_{AA} + \Delta_A \Delta_A^\top)^{-1}\|_2 \left(1 + \mathcal{O}\left(\sqrt{\frac{s \log(\log(n))}{n}}\right)\right).$$

$\square$

**Lemma 12.** *With probability at least $1 - \log^{-1}(n)$*

$$\|D_A^\top \Sigma_{AA}^{-1} D_A - D_A^\top W_{AA}^{-1} D_A\|_2 \leq C \|D_A^\top \Sigma_{AA}^{-1} D_A\|_2 \sqrt{\frac{(G-1) \log(\log(n))}{n}}.$$

*Proof of Lemma 12.* By submultiplicity of operator norm,

$$\|D_A^\top \Sigma_{AA}^{-1} D_A - D_A^\top W_{AA}^{-1} D_A\|_2 \leq \|D_A^\top \Sigma_{AA}^{-1} D_A\|_2 \|I - (D_A^\top \Sigma_{AA}^{-1} D_A)^{-1/2} D_A^\top W_{AA}^{-1} D_A (D_A^\top \Sigma_{AA}^{-1} D_A)^{-1/2}\|_2.$$

By Theorem 3.2.5 and Theorem 3.2.11 in [15],

$$(n - G)(D_A^\top \Sigma_{AA}^{-1} D_A)^{1/2} (D_A^\top W_{AA}^{-1} D_A)^{-1} (D_A^\top \Sigma_{AA}^{-1} D_A)^{1/2} \sim W_{G-1}(n - s - 1, I).$$

By Lemma 9 in [20], with probability at most $2 \exp\left(-(n - s - 1)t^2/2\right)$,

$$\|\frac{n - s - 1}{n - G}(D_A^\top \Sigma_{AA}^{-1} D_A)^{-1/2} D_A^\top W_{AA}^{-1} D_A (D_A^\top \Sigma_{AA}^{-1} D_A)^{-1/2} - I\|_2 \geq \delta(n - s - 1, G - 1, t),$$

where

$$\delta(n - s - 1, G - 1, t) = 2 \left(\sqrt{\frac{G-1}{n-s-1}} + t\right) + \left(\sqrt{\frac{G-1}{n-s-1}} + t\right)^2.$$

Let
$$t = \sqrt{\frac{2\log(2\log n))}{n-s-1}}.$$

Then with probability at least $1 - \log^{-1}(n)$

$$\|\frac{n-s-1}{n-G}(D_A^\top \Sigma_{AA}^{-1} D_A)^{-1/2} D_A^\top W_{AA}^{-1} D_A (D_A^\top \Sigma_{AA}^{-1} D_A)^{-1/2} - I\|_2 \le 8\sqrt{\frac{2(G-1)\log(2\log(n))}{n-s-1}}.$$

Hence, with probability at least $1 - \log^{-1}(n)$

$$\|(D_A^\top \Sigma_{AA}^{-1} D_A)^{-1/2} D_A^\top W_{AA}^{-1} D_A (D_A^\top \Sigma_{AA}^{-1} D_A)^{-1/2} - I\|_2 \le C\sqrt{\frac{(G-1)\log(\log(n))}{n}}.$$

$\square$

**Lemma 13.** *With probability at least $1 - \log^{-1}(n)$*

$$\|D_A^\top \Sigma_{AA}^{-1} D_A\|_2 \le (G-1)\|\Delta_A^\top \Sigma_{AA}^{-1} \Delta_A\|_2$$
$$+ \mathcal{O}\left(\frac{(G-1)s\log(\log(n))}{n} \vee \sqrt{\|\Delta_A^\top \Sigma_{AA}^{-1} \Delta_A\|_2 \frac{(G-1)\log(\log(n))}{n}}\right).$$

*Proof of Lemma 13.* Since $D_A^\top \Sigma_{AA}^{-1} D_A$ is a positive semi-definite matrix,

$$\|D_A^\top \Sigma_{AA}^{-1} D_A\|_2 \le \operatorname{Tr}(D_A^\top \Sigma_{AA}^{-1} D_A).$$

Recall that $D_A \sim \mathcal{N}(\Delta_A, \Sigma_{AA}/n \otimes I)$. Therefore for all $i \in \{1,..,(G-1)\}$

$$ne_i^\top D_A^\top \Sigma_{AA}^{-1} D_A e_i \sim \chi_s^2\left(ne_i^\top \Delta_A^\top \Sigma_{AA}^{-1} \Delta_A e_i\right).$$

From [9, Lemma 11], with probability at least $1 - \log^{-1}(n)$, for all $i \in \{1,..,(G-1)\}$

$$e_i^\top D_A^\top \Sigma_{AA}^{-1} D_A e_i \le e_i^\top \Delta_A^\top \Sigma_{AA}^{-1} \Delta_A e_i$$
$$+ \mathcal{O}\left(\frac{s\log((G-1)\log(n))}{n} \vee \sqrt{e_i^\top \Delta_A^\top \Sigma_{AA}^{-1} \Delta_A e_i \frac{\log((G-1)\log(n))}{n}}\right),$$

or equivalently

$$\operatorname{Tr}(D_A^\top \Sigma_{AA}^{-1} D_A) \le \operatorname{Tr}(\Delta_A^\top \Sigma_{AA}^{-1} \Delta_A)$$
$$+ \mathcal{O}\left(\frac{(G-1)s\log((G-1)\log(n))}{n} \vee \sqrt{\operatorname{Tr}(\Delta_A^\top \Sigma_{AA}^{-1} \Delta_A) \frac{\log((G-1)\log(n))}{n}}\right).$$

Since $\operatorname{Tr}(\Delta_A^\top \Sigma_{AA}^{-1} \Delta_A) \le (G-1)\|\Delta_A^\top \Sigma_{AA}^{-1} \Delta_A\|_2$ and $G = \mathcal{O}(1)$, it follows that with probability at least $1 - \log^{-1}(n)$

$$\|D_A^\top \Sigma_{AA}^{-1} D_A\|_2 \le (G-1)\|\Delta_A^\top \Sigma_{AA}^{-1} \Delta_A\|_2$$
$$+ \mathcal{O}\left(\frac{(G-1)s\log(\log(n))}{n} \vee \sqrt{\|\Delta_A^\top \Sigma_{AA}^{-1} \Delta_A\|_2 \frac{(G-1)\log(\log(n))}{n}}\right).$$

$\square$

**Lemma 14.** *With probability at least $1 - \mathcal{O}(\log^{-1}(n))$*

$$\|D_A^\top W_{AA}^{-1} D_A\|_2 \leq C\|\Delta_A^\top \Sigma_{AA}^{-1} \Delta_A\|_2 + \mathcal{O}\left(\frac{(G-1)s\log(\log(n))}{n} \vee \sqrt{\|\Delta_A^\top \Sigma_{AA}^{-1} \Delta_A\|_2 \frac{(G-1)\log(\log(n))}{n}}\right).$$

*Proof of Lemma 14.* By triangle inequality and Lemma 13,

$$\begin{aligned}
\|D_A^\top W_{AA}^{-1} D_A\|_2 &= \frac{\|D_A^\top W_{AA}^{-1} D_A\|_2}{\|D_A^\top \Sigma_{AA}^{-1} D_A\|_2} \|D_A^\top \Sigma_{AA}^{-1} D_A\|_2 \\
&\leq \frac{\|D_A^\top W_{AA}^{-1} D_A\|_2}{\|D_A^\top \Sigma_{AA}^{-1} D_A\|_2} \Big((G-1)\|\Delta_A^\top \Sigma_{AA}^{-1} \Delta_A\|_2 \\
&\quad + \mathcal{O}\left(\frac{(G-1)s\log(\log(n))}{n} \vee \sqrt{\|\Delta_A^\top \Sigma_{AA}^{-1} \Delta_A\|_2 \frac{(G-1)\log(\log(n))}{n}}\right)\Big).
\end{aligned}$$

From Lemma 12, with probability at least $1 - \log^{-1}(n)$

$$\begin{aligned}
\frac{\|D_A^\top W_{AA}^{-1} D_A\|_2}{\|D_A^\top \Sigma_{AA}^{-1} D_A\|_2} &\leq \frac{\|D_A^\top \Sigma_{AA}^{-1} D_A\|_2 + \|D_A^\top W_{AA}^{-1} D_A - D_A^\top \Sigma_{AA}^{-1} D_A\|_2}{\|D_A^\top \Sigma_{AA}^{-1} D_A\|_2} \\
&\leq 1 + C\sqrt{\frac{(G-1)\log(\log(n))}{n}} \\
&\leq C'.
\end{aligned}$$

Combining with the previous display, we obtain with probability at least $1 - \mathcal{O}(\log^{-1}(n))$ that

$$\begin{aligned}
\|D_A^\top W_{AA}^{-1} D_A\|_2 &\leq C\|\Delta_A^\top \Sigma_{AA}^{-1} \Delta_A\|_2 \\
&\quad + \mathcal{O}\left(\frac{(G-1)s\log(\log(n))}{n} \vee \sqrt{\|\Delta_A^\top \Sigma_{AA}^{-1} \Delta_A\|_2 \frac{(G-1)\log(\log(n))}{n}}\right).
\end{aligned}$$

□